*Article*

# UKANFormer: Noise-Robust Semantic Segmentation for Coral Reef Mapping via a Kolmogorov–Arnold Network-Transformer Hybrid

**Tianyang Dou [1], Ming Li [2,\*], Jiangying Qin [3], Xuan Liao [4], Jiageng Zhong [5], Armin Gruen [6], Mengyi Deng [7]**

[1] School of Remote Sensing and Information Engineering, Wuhan University, 430079 Wuhan, China; tianyangdou@whu.edu.cn
[2] School of Robotics, Wuhan University, 430079 Wuhan, China; lisouming@whu.edu.cn
[3] State Key Laboratory of Information Engineering in Surveying, Mapping and Remote Sensing, Wuhan University, 430079 Wuhan, China; jy_qin@whu.edu.cn
[4] Department of Land Surveying and Geo-Informatics, The Hong Kong Polytechnic University, Hong Kong, China; xuan-lsgi.liao@connect.polyu.hk
[5] State Key Laboratory of Information Engineering in Surveying, Mapping and Remote Sensing, Wuhan University, 430079 Wuhan, China; zhongjiageng@whu.edu.cn
[6] Institute of Geodesy and Photogrammetry, ETH Zürich, 8093 Zürich, Switzerland; armin.gruen@geod.baug.ethz.ch
[7] Red Cross Society of China Honghu Branch, 433200 Jingzhou, China;
\* Correspondence: lisouming@whu.edu.cn

**Highlights**

- We propose UKANFormer, a novel segmentation model that integrates Kolmogorov–Arnold networks (KAN) and a transformer-based decoder (GL-Trans) for robust satellite coral image segmentation.
- UKANFormer achieves 67.00% coral-class IoU and 83.98% pixel accuracy when evaluated against expert-annotated coral labels, outperforming conventional baselines.
- Our method enables high-resolution, globally scalable coral reef mapping under noisy supervision, offering a practical solution for global ecological monitoring.

**Abstract**

Coral reefs are vital yet fragile ecosystems that require accurate large-scale mapping for effective conservation. Although global products such as the Allen Coral Atlas provide unprecedented coverage of global coral reef distribution, their predictions are frequently limited in spatial precision and semantic consistency, especially in regions requiring fine-grained boundary delineation. To address these challenges, we propose UKANFormer, a novel semantic segmentation model designed to achieve high-precision mapping under noisy supervision derived from Allen Coral Atlas. Building upon the UKAN architecture, UKANFormer incorporates a Global-Local Transformer (GL-Trans) block in the decoder, enabling the extraction of both global semantic structures and local boundary details. In experiments, UKANFormer achieved a coral-class IoU of 67.00% and pixel accuracy of 83.98%, outperforming conventional baselines under the same noisy labels setting. Remarkably, the model produces predictions that are visually and structurally more accurate than the noisy labels used for training. These results challenge the notion that data quality directly limits model performance, showing that architectural design can mitigate label noise and support scalable mapping under imperfect supervision. UKANFormer provides a foundation for ecological monitoring where reliable labels are scarce.

## 1. Introduction

The Coral reefs are ecologically critical marine ecosystems, supporting a large share of ocean biodiversity[1,2]. However, they are rapidly degrading due to global warming[3], marine pollution[4], and other human pressures. Recent assessments indicate that the world has already lost about 1.4% of coral reefs since 2009, and under high-emission warming scenarios, up to 99% could undergo bleaching by the end of this century[5-7].

In this context, accurate and up-to-date mapping of coral reef habitats is essential for large-scale ecological monitoring and conservation planning[8]. Yet, the most reliable mapping methods—such as underwater photogrammetry and diver-based surveys—are expensive, labor-intensive, and geographically constrained[9,10]. Satellite images offers a scalable alternative, enabling frequent, wide-area monitoring of coral reef systems[11,12].

One of the most widely used resources for global coral reef monitoring is the Allen Coral Atlas[13,14], which provides 5-meter resolution maps of coral reef geomorphology and benthic habitats based on Planet-5 and Sentinel-2 imagery. While the Allen dataset represents a major step forward in standardized coral reef mapping, its map products were generated through rule-based classification and machine learning techniques[15,16]. As a result, the outcomes often exhibit imprecision, especially in heterogeneous coral reef environments, limiting their utility for fine-grained analysis. However, despite its limitations, the Allen Coral Atlas remains the few publicly available global-scale dataset providing standardized coral labels suitable for large-scale model training. Consequently, any data-driven approach to coral segmentation must be developed under the constraints of limited precision and label noise inherent in the Allen dataset.

This leads to a fundamental challenge for the development of data-driven segmentation models: how can we build accurate, high-resolution coral reef mapping models when the available training labels are themselves noisy and coarse? While conventional wisdom emphasizes the need for better data, we explore an alternative route—can better models compensate for imperfect supervision?

In this study, we address this challenge by proposing UKANFormer, a semantic segmentation model designed to produce robust and fine-grained coral reef predictions from supervision with noisy and coarse labels. Building on the UKAN architecture, UKANFormer introduces a Global-Local Transformer (GL-Trans) block in the decoder. This block combines local convolutions, which enhance spatial detail and boundary precision, with global self-attention mechanisms that capture large-scale contextual dependencies and semantic coherence, allowing the model to extract both intricate local patterns and overarching structural information from satellite images. Importantly, this architectural design directly responds to the dual challenge of noisy labels and complex spatial structures in coral reef imagery. To evaluate UKANFormer's effectiveness under noisy supervision, we constructed a high-quality test set of coral reef image patches, annotated by experts in remote sensing and ecology. In our experiments, despite being trained exclusively on coarse labels, UKANFormer consistently demonstrates strong robustness to label noise, often producing predictions that implicitly correct flawed supervision.

Overall, this study contributes a noise-robust segmentation framework and a curated expert-labeled benchmark for evaluating coral mapping models under imperfect supervision.

This paper is structured as follows: Section 2 reviews related work on coral reef mapping datasets and the evolution of semantic segmentation technologies. Section 3 introduces the architecture of UKANFormer and the GL-Trans block. Section 4 details the dataset preparation, implementation, and evaluation protocols. Section 5 presents a discussion of our findings, their broader implications, and the study's limitations.

## 2. Related Work

*2.1. Allen Dataset*

The Allen Coral Atlas dataset, released by Lyon et al.[13] in 2024, is one of the most widely used global coral reef satellite datasets. It provides 5-meter high-resolution global maps of coral reef geomorphology and benthic substrates, constructed from over 1.17 million Planet-5 Dove[17,18] satellite images and 1.05 million Sentinel-2[19] scenes collected between 2018 and 2020.For training, the project integrated over 1.5 million labeled samples from more than 480 collaborators, including approximately 500,000 in situ photo quadrats, covering 11 geomorphic classes and 6 benthic substrates. The mapping pipeline included: (1) multi-source image stack construction, (2) training of a Random Forest classifier with about 2,000 samples per class, and (3) rule-based post-processing with manual refinement. The process was implemented in parallel across 30 global coral reef regions via Google Earth Engine. A key strength of the Allen dataset lies in its open-access availability and wide applicability. It has been used in marine spatial planning, blue carbon estimation, and policy development. Accuracy assessment, based on ~78,000 geomorphic and ~57,000 benthic validation points, reported average accuracies of 69% (geomorphic) and 66% (benthic), varying across regions.

It should be noted that this mapping effort lacks in situ validation in our study areas (these coral islands in the South China Sea). Moreover, because the satellite imagery was acquired from different sensors and at different times, large-scale satellite observations are inevitably affected by adverse weather conditions such as cloud cover and rainfall. Consequently, the segmentation accuracy in these regions may be even lower. Nevertheless, while we acknowledge the significance of this work as the first global shallow-water coral reef mapping product based on satellite observations, there remains considerable scope for further research to enhance the accuracy of global coral reef mapping derived from satellite and other remote sensing data.

*2.2. The Development of Semantic Segmentation Technologies*

Semantic segmentation is a fundamental task in satellite image analysis, aiming to assign semantic labels to every pixel in an image. While early methods—such as Full Convolutional Network (FCN)[20], UNet[21], and the DeepLab series[22-25]—demonstrated strong performance in structured environments, their limited receptive fields restrict their ability to model long-range dependencies, making them less effective in spatially less effective in spatially heterogeneous scenes like coral reefs.

To address this limitation, Transformer-based architectures, such as SegFormer[26] and Swin Transformer[27,28], were introduced to overcome this limitation by capturing global context via self-attention. These models have achieved state-of-the-art results on various remote sensing benchmarks. However, their high computational cost and lack of interpretability present challenges for scalable, explainable ecological monitoring[29].

In response to these challenges, the Kolmogorov–Arnold Network (KAN)[30] architecture was proposed, grounded in the Kolmogorov–Arnold representation theorem[31,32]. KAN shifts nonlinearity from activation functions at nodes to learnable spline-based functions on the network's edges, leading to smoother function approximation and improved generalization. The UKAN[33] architecture further extends KANs into vision tasks by introducing a modular U-shaped design with enhanced capability for detail modeling and cross domain transfer.

Building upon this, we propose UKANFormer, a hybrid architecture that combines a KAN-based encoder with a Global-Local Transformer(GL-Trans) decoder. This design integrates local convolution for details preservation and noise-tolerant solution for coral segmentation under large-scale and imperfect labeling.

## 3. Methodology

UKANFormer enhances the original UKAN architecture by introducing a Transformer-based decoder (the GL-Trans block), thereby addressing its limitations in global context modeling. This extension was motivated by empirical observations that UKAN struggled with spatial continuity and semantic coherence in complex coral reef scenes. With its modular structure, UKANFormer also demonstrates strong scalability and generalization potential, making it well-suited for semantic segmentation tasks across diverse ecological settings and multi-source satellite imagery.

*3.1. UKAN: Baseline Architecture*

UKAN (U-shaped Kolmogorov–Arnold Network) is a semantic segmentation framework designed to enhance nonlinear modeling capacity and robustness under label noise. It is based on the Kolmogorov-Arnold representation theorem, which enables network to approximate multivariate functions through combinations of univariate nonlinear mappings. This design makes it particularly suitable for tasks with subtle transitions and ambiguous boundaries[34], such as coral segmentation.

The architecture consists of a convolutional encoder, a tokenized KAN(Tok-KAN) module, and a decoder with skip connections. The encoder extracts low-level visual features from the input image using standard CNN blocks. These features are then partitioned into small patches and then projected into a lower-dimensional embedding space, serving as input tokens to the KAN module.

Unlike conventional multilayer perceptrons (MLPs), the KAN layers within Tok-KAN replace fixed weight matrices with learnable activation functions composed of basic functions. This unique design enables each layer to perform flexible, data-adaptive nonlinear mappings rather than static linear transformations followed by predefined activations. By recursively composing such mappings, the KAN layers effectively form a deep functional transformation tailored to the input structure[35,36].

This architectural design allows UKAN to learn smooth yet expressive mappings, capturing fine-grained semantic details even in the presence of imperfect annotations. By combining the tokenized patch embeddings with the hierarchical nonlinear transformations of KAN layers, UKAN captures both local and global semantic patterns[37,38]. The transformed feature tokens are finally forwarded to the decoding stage for semantic segmentation.

In our framework, UKAN serves as a strong and extensible baseline model. Establishing on this foundation, we introduce a global-context-aware decoding block that builds upon this foundation to further improve spatial consistency and semantic coherence in challenging coral segmentation scenarios.

**Figure 1** illustrates the overall architecture of the UKAN baseline model, which includes the convolutional encoder, the Tok-KAN module, and the decoder structure. **Figure 2** presents a detailed view of the KAN layer's structure and the flow of information within it, providing a clearer understanding of the role each activation function plays in the KAN layer.

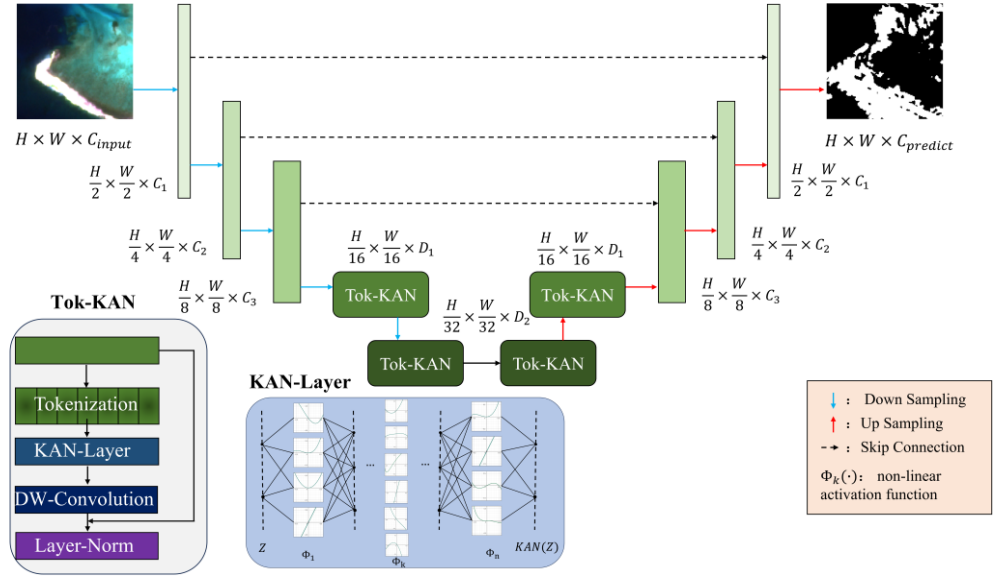

**Figure 1. Overview of the UKAN Baseline Architecture**
The UKAN model consists of a CNN-based encoder, a Tok-KAN module for nonlinear transformation, and a decoder that integrates local and global semantic information. Skip connections are used for multi-scale feature reuse to ensure accurate segmentation.

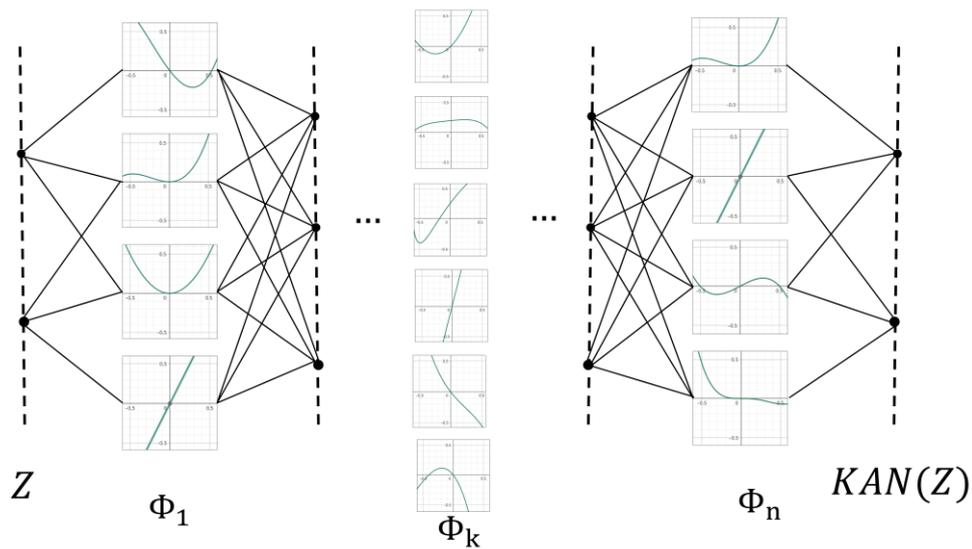

**Figure 2. KAN Layer Structure and Information Flow**
This figure illustrates the internal structure of the KAN layer, highlighting the recursive nature of the layer-wise transformations. Each layer of the KAN module consists of multiple learnable activation functions, which are combined to form the overall nonlinear transformation. The output of one layer serves as the input for the subsequent layer, enabling the KAN network to capture complex, hierarchical features in the input data.

*3.2. GL-Trans Block*

To mitigate the structural inconsistencies often found in coarse coral labels—such as semantic fragmentation and global discontinuity—we propose the GL-Trans block, a novel decoder block that enhances long-range structure modeling under noisy supervision. Inspired by the dual-branch attention architecture proposed by Wang et al[29]., it is integrated into the decoder of our UKANFormer model to address a key limitation in the original UKAN: its insufficient ability to capture the global structure and connectivity of coral regions. The architecture of GL-Trans is illustrated in **Figure 3**, consisting of a Local Branch and a Global Branch.

3.2.1. The Local Branch

In the Local Branch, the input feature tensor $X \in \mathbb{R}^{C \times H \times W}$ is first represented as a 2D feature map, where $C$, $H$ and $W$ represent the number of channels, height and width, respectively. It then passes through two parallel convolutional paths. The first path employs a $1 \times 1$ convolution to perform channel-wise integration and transformation, formulated as:

$$F_{1\times1}(X) = \text{BatchNorm}(W_{1\times1} * X) \tag{1}$$

Batch normalization stabilizes training and improves generalization. The second path uses a $3 \times 3$ convolution to capture local spatial textures and edge information:

$$F_{3\times3}(X) = \text{BatchNorm}(W_{3\times3} * X) \tag{2}$$

The outputs of these two paths are then summed to obtain a comprehensive local contextual representation:

$$F_{local} = F_{1\times1}(X) + F_{3\times3}(X) \tag{3}$$

3.2.2. The Global Branch

The Global Branch, inspired by the Transformer self-attention mechanism, first reshapes the input feature tensor $X \in \mathbb{R}^{C \times H \times W}$ into a sequence form $X_{seq} \in \mathbb{R}^{L \times D}$, where the sequence length $L = H \times W$ and $D$ is the feature dimension of each token after a linear projection[39]. This dimension $D$ represents the number of semantic channels used to encode each spatial location, and is typically configurable, independent of the original input channel size $C$. This sequence is linearly projected into query ($Q$), key ($K$), and value ($V$) matrices as follows:

$$\begin{cases} Q = W_Q \cdot X_{seq} \\ K = W_K \cdot X_{seq} \\ V = W_V \cdot X_{seq} \end{cases} \tag{4}$$

where $W_Q, W_K, W_V \in \mathbb{R}^{D \times D_k}$ are learnable weight matrices. The attention weights are computed using the scaled dot-product attention, augmented by a learnable bias $\theta \in \mathbb{R}^{L \times L}$

$$A = \text{Softmax}\left(\frac{Q \cdot K^{\text{T}}}{\sqrt{D_k}} + \theta\right) \tag{5}$$

which effectively captures long-range dependencies and semantic relationships between features. The weighted sum of values is then reshaped back to the spatial feature map form to produce the global contextual feature:

$$F_{global} = \text{Reshape}(AV) \tag{6}$$

After fusing the local and global features, a depth-wise separable convolution is applied for further refinement, which reduces computational cost while maintaining representational power:

$$F_{Dw} = \text{DWConv}(F_{local} + F_{global}) \tag{7}$$

The block output is obtained via a 1×1 convolution and then a batch normalization.

$$F_{out} = \text{BatchNorm}(W_{1\times1} * F_{Dw}) \tag{8}$$

Through this multi-path and multi-scale fusion mechanism, the GL-Trans block achieves synchronous perception of fine-grained textures and long-range semantics in satellite images, significantly improving segmentation performance and robustness[40]. This design is particularly beneficial when training on imperfect labels, where reinforcing structural consistency helps mitigate the adverse effects of label noise.

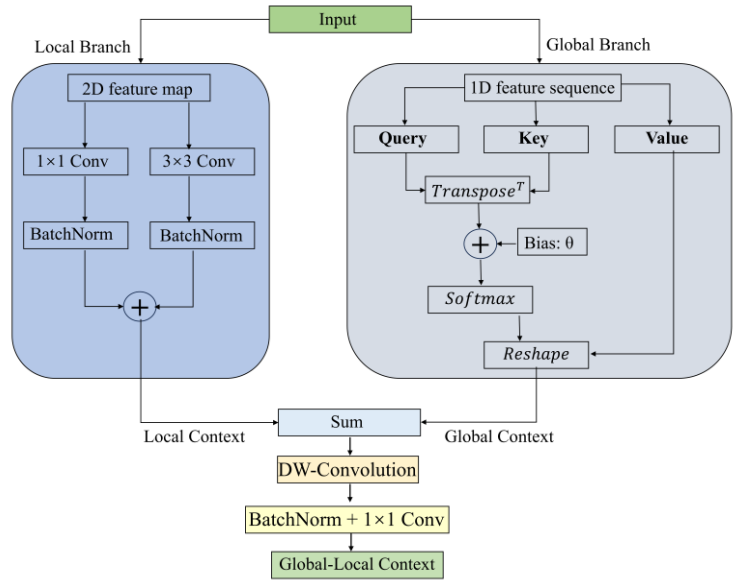

**Figure 3. Architecture of the GL-Trans block.**
A dual-branch block that fuses local convolutional features with global self-attention, enhancing both spatial detail and semantic coherence in coral semantic segmentation tasks**.**

*3.3. UKANFormer*

UKANFormer is a novel encoder–decoder architecture tailored for semantic segmentation of satellite imagery under coarse and noisy supervision. As illustrated in **Figure 4**, the model comprises three major components: a CNN-based backbone encoder, a Tok-KAN block grounded in Kolmogorov–Arnold Networks (KAN), and a GL-Trans decoder that integrates both local and global contextual features. These modules operate across multiple spatial scales to jointly capture fine-grained details and holistic structure information.

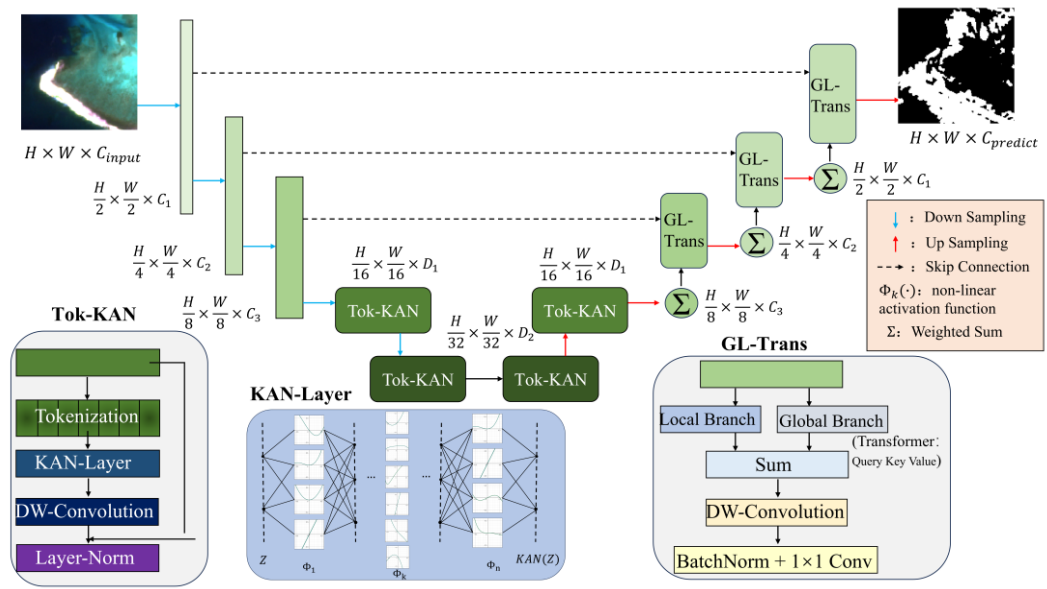

**Figure 4. Overall architecture of the proposed UKANFormer model.**
The overall architecture consists of a CNN-based encoder for hierarchical feature extraction, a Tok-KAN module that enhances non-linear modeling capacity and boundary detail representation, and a GL-Trans decoder that fuses local and global semantic information to produce coherent segmentation maps. Skip connections are incorporated to facilitate feature reuse across multiple spatial scales and to preserve fine structural details.

In the encoding stage, low-level visual features are first extracted using a convolutional backbone, which progressively down-samples the input image into hierarchical feature maps. The Tok-KAN block, introduced after the encoder, is based on the Kolmogorov–Arnold Network, which was chosen for its ability to enhance non-linear modeling and interpretability — key advantages in handling subtle coral textures and label noise.

The decoding stage is handled by the GL-Trans block, which incorporates a dual-path structure combining a local convolutional branch and a global transformer branch. The global path captures long-range dependencies and ensures semantic coherence across spatial regions, while the local path retains high-frequency details and boundary contours. These two branches are fused through attention-based aggregation and are progressively up-sampled with skip connections to reconstruct full-resolution segmentation maps.

## 4. Data and Experiments

*4.1. Study Areas and data Preprocessing*

This study focuses on two ecologically and geographically distinct coral reef regions: the South China Sea and French Polynesia. Within the South China Sea, we selected four representative coral reef systems—Yongxing Island, Huaguang Atoll, Yuzhuo Atoll, and Langhua Atoll—while in French Polynesia, the selected sites included Moorea Island, Tetiaroa Atoll, and Tahiti Island. The geographic extent of each coral reef site, including latitude and longitude bounds, is summarized in **Table 1**. These coral reefs vary in geomorphological structure, benthic composition, and environmental conditions, making them representative of a wide range of tropical coral reef types and suitable for evaluating the generalization ability of semantic segmentation models.

The satellite imagery used in this study was sourced from the Planet-5(Planet-Scope) satellite constellation, specifically the PSScene sensor, with a spatial resolution of 3 meters. To ensure spectral consistency across scenes and enhance comparability, we used the Harmonized version of Planet-Scope products. All images were orthorectified and reprojected into the WGS-84 coordinate reference system, ensuring spatial consistency across regions.

Preprocessing began with spatial clipping of each coral reef scene based on its geographical extent. To promote edge continuity and improve patch-level learning, we applied a 20% overlap in both horizontal and vertical directions during image tiling. The scenes were then subdivided into 256×256-pixel patches, yielding a total of 1388 remote sensing patches. From this dataset, approximately 10% (138 patches) were randomly selected to serve as the test set and were annotated by domain specialists with expertise in coral reef ecosystems. These samples include various benthic patterns and coral reef distributions to ensure diversity. The remaining patches were used exclusively for model training.

Supervision labels were derived from the Allen Coral Atlas benthic habitat maps, which provide categorical benthic classification for all seven coral reef regions. For the purposes of this study, we focused specifically on coral habitat identification and reclassified the maps into a binary scheme: pixels labeled as "coral" were treated as the positive class, and all other benthic types were grouped as the negative class. All label masks were geometrically aligned with the satellite imagery to enable precise pixel-wise training. **Figures 5 and 6** present the spatial distribution of the selected coral reef sites in the South China Sea and French Polynesia, respectively.

| Region | Island/Atoll Name | Latitude Range | Longitude Range |
|---|---|---|---|
| *South China Sea* | Yongxing Island | 16.79°N ~ 16.85°N | 112.32°E ~ 112.40°E |
| | Huaguang Atoll | 16.13°N ~ 16.28°N | 111.52°E ~ 111.84°E |
| | Yuzhuo Atoll | 16.30°N ~ 16.38°N | 111.94°E ~ 112.10°E |
| | Langhua Atoll | 15.99°N ~ 16.11°N | 112.43°E ~ 112.61°E |
| *French Polynesia* | Moorea Island | 17.43°S ~ 17.68°S | 149.67°W ~ 150.00°W |
| | Tahiti Island | 17.42°S ~ 17.97°S | 149.04°W ~ 149.70°W |
| | Tetiaroa Atoll | 16.97°S ~ 17.06°S | 149.51°W ~ 149.62°W |

**Table 1. Geographic extent of study coral reef sites.**

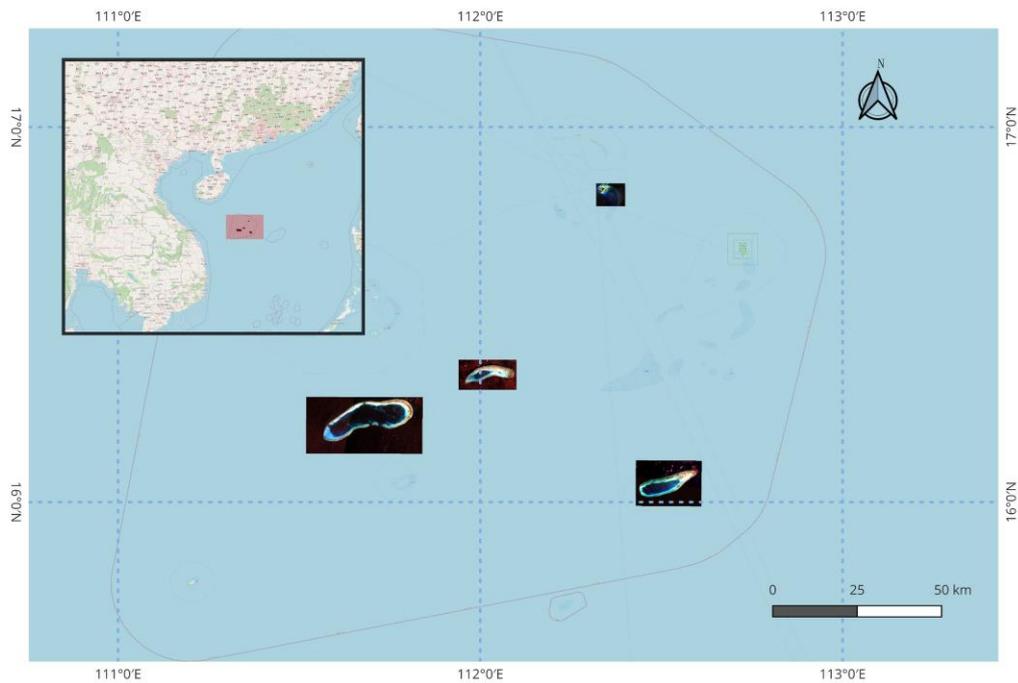

**Figure 5. Geographic locations of selected coral reef sites in the South China Sea.**
The map shows the positions of Yongxing Island, Huaguang Atoll, Yuzhuo Atoll, and Langhua Atoll, representing typical tropical coral reef systems in the South China Sea.

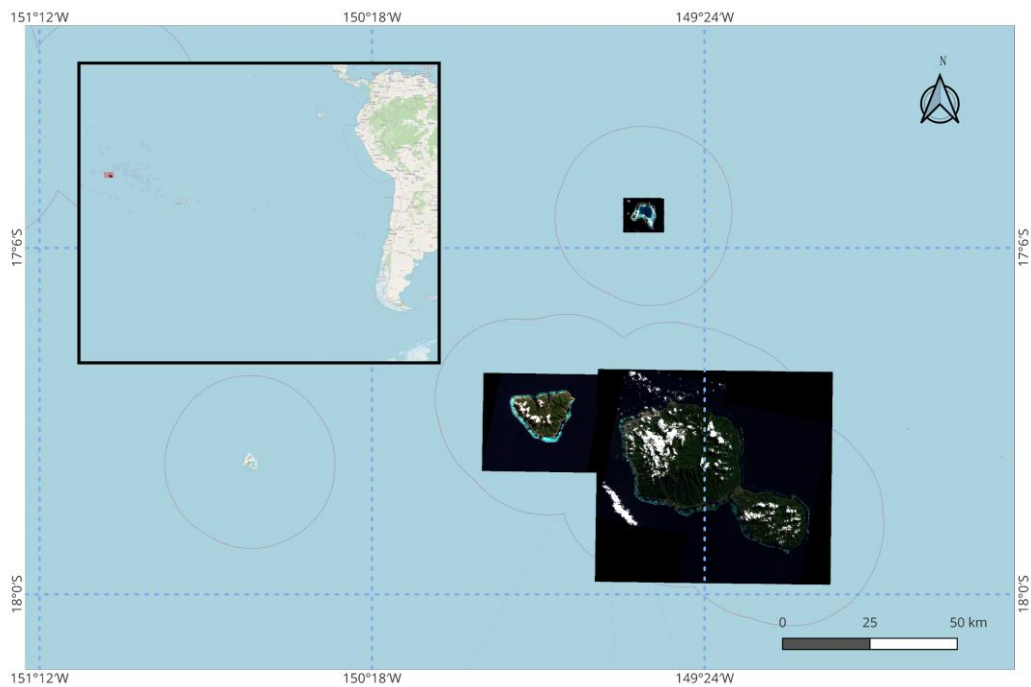

**Figure 6. Geographic locations of selected coral reef sites in French Polynesia.**
The map displays the locations of Moorea Island, Tetiaroa Atoll, and Tahiti Island, illustrating representative coral reef types in the central Pacific Ocean.

*4.2. Implemented Details*

All models were implemented using PyTorch 2.5.1 with CUDA 11.8 and trained on a single NVIDIA RTX 4070 GPU. We set the batch size to 8 and trained the models for 300 epochs using the Stochastic Gradient Descent (SGD) optimizer. The initial learning rate was set to $2.5\times10^{-6}$ and adjusted dynamically using the ReduceLROnPlateau scheduler. The cross entropy loss was used as the objective function. During training, data augmentation techniques including random horizontal flipping and mirroring were applied. Input images were resized to 256×256 pixels as described in the data preprocessing section.

*4.3. Comparison between UKANFormer and SOTA Methods*

To evaluate the segmentation accuracy and robustness of UKANFormer, we compare its performance with three representative state-of-the-art baselines introduced in Section 2: UKAN[33], UNetFormer[29], and UNet[21]. Two distinct label standards were used for evaluation: expert-annotated high-quality coral labels and the noisy, coarse Allen dataset labels. **Table 2** reports the Intersection over Union (IoU) of coral class and pixel accuracy achieved by each model on the coral class, while **Figure 7** provides a visual comparison of the predictions on the same image patch to support both quantitative and qualitative analysis.

4.3.1. Quantitative Results Analysis

As shown in **Table 2**, UKANFormer consistently outperforms all baseline models across both labeling standards, demonstrating its robustness to annotation noise and its superior capability in modeling coral structures. When evaluated against our expert-annotated labels, UKANFormer achieves the highest performance, followed by UKAN, UNet, and UNetFormer. The performance gap is especially evident in structurally complex or transition-prone reef areas, where UKANFormer better preserves spatial continuity and delineates fine-grained boundaries.

Under the Allen dataset labels—known for their coarse quality and boundary ambiguity—all models exhibit a performance drop, as expected. Despite this, UKANFormer remains the top performer, indicating its strong generalization capacity even when trained or evaluated with noisy annotations.

It is important to note that the Allen Coral Atlas primarily emphasizes pixel-wise accuracy as the main quality metric, reflecting its application focus on large-scale reef monitoring. Pixel accuracy, defined as the proportion of correctly classified pixels across the entire image, tends to yield higher scores even when segmentation boundaries are imprecise. In contrast, IoU measures the per-image overlap between predicted and reference coral regions, calculated as the intersection over union of the coral class in each image and then averaged across the test set. It is inherently more sensitive to boundary shifts and fragmented predictions. This distinction explains why pixel accuracy scores are generally higher than coral-class IoU across all models and labeling standards. Accuracy is less sensitive to spatial structure and class imbalance, while IoU penalizes small misalignments, missing regions, and over-segmented areas more heavily.

Furthermore, we observe that the performance drop from expert labels to Allen labels is more pronounced in coral-class IoU than in pixel accuracy (i.e., Δ IoU > Δ Accuracy across most models). This suggests that label coarseness and boundary ambiguity disproportionately affect structure-aware metrics. Given the inherently irregular and fragmented morphology of coral reefs, IoU provides a more informative assessment of segmentation quality, particularly in applications requiring detailed reef mapping and spatial consistency.

Together, these results suggest that UKANFormer not only aligns well with expert interpretations of coral distribution, but also exhibits robustness when evaluated against less reliable annotations, making it a promising candidate for scalable reef mapping in data-constrained scenarios.

| Model | Reference Labels | Coral-class IoU(%) | Accuracy(%) |
| --- | --- | --- | --- |
| UKANFormer | Our Labels | **67.00** | **83.98** |
| | Allen Labels | 47.67 | 66.49 |
| UKAN | Our Labels | 64.58 | 81.88 |
| | Allen Labels | 45.56 | 65.47 |
| UNetFormer | Our Labels | 53.95 | 72.30 |
| | Allen Labels | 40.67 | 63.89 |
| UNet | Our Labels | 64.25 | 76.88 |
| | Allen Labels | 43.05 | 63.68 |

**Table 2. Quantitative Results Analysis**

Performance comparison of different models on coral semantic segmentation under two annotation standards: our expert-annotated labels and Allen dataset labels. Metrics include coral-class Intersection over Union (IoU) and pixel accuracy (%). UKANFormer consistently outperforms baseline models across both standards.

4.3.2. Qualitative Results Analysis

**Figure 7** illustrates the qualitative performance of the compared models. UKANFormer produces segmentation maps with sharp boundaries and continuous coral regions, even under challenging imaging conditions.

Compared with UKAN—which preserves local textures but tends to lose global consistency—UKANFormer yields smoother and more coherent predictions, particularly along region boundaries, due to its GL-Trans block that enhances global semantic modeling.

In contrast, the traditional UNet model exhibits noticeable block-like artifacts, especially near coral edges. These artifacts can be attributed to the complexity of coral boundaries in satellite imagery, which are often blurred and irregular due to factors such as lighting, water depth, and surface reflection—patterns that shallow convolutional layers struggle to capture. Moreover, UNet's upsampling relies on transposed convolutions or interpolation with limited skip connections, lacking global context integration, which leads to spatial misalignments during feature reconstruction.

By incorporating multi-scale feature aggregation and a Transformer-based global modeling block, UKANFormer overcomes these limitations. It unifies local detail preservation and global semantic consistency, resulting in structurally coherent and visually consistent coral segmentation outputs.

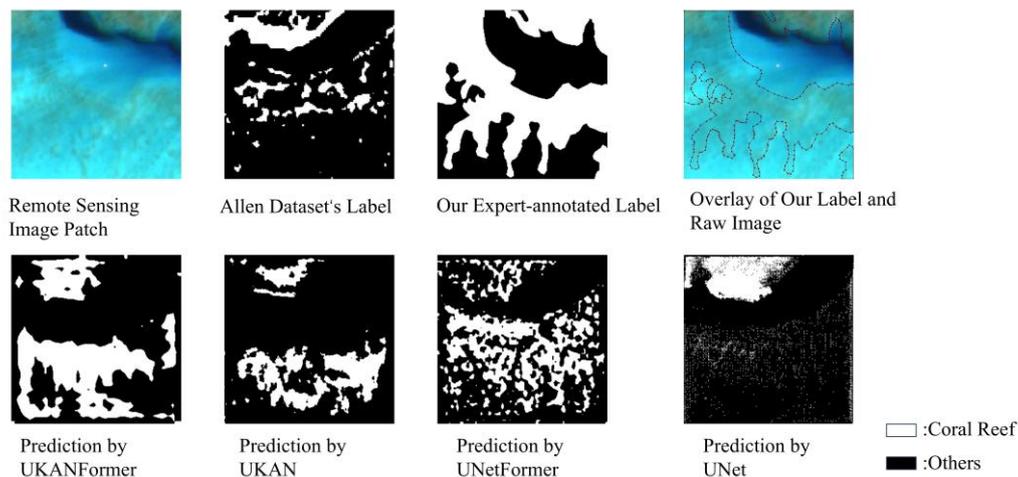

**Figure 7. Qualitative Results Analysis**

Visual comparison of segmentation results from UKANFormer, UKAN, UNetFormer, and UNet on the same satellite image patch.

4.3.3. Ablation Study Perspective: UKANFormer vs. UKAN

To further isolate the contribution of the GL-Trans block, we conducted an ablation-style comparison between UKANFormer and its baseline architecture UKAN.

UKANFormer differs from UKAN primarily by incorporating the GL-Trans block into the decoder, which fuses local convolutional features with global self-attention to strengthen structural consistency. As reported in Table 2, this addition improves the coral-class IoU by 2.42 percentage points and the pixel accuracy by 2.10 percentage points under expert-labeled evaluation. Comparable gains of approximately 2 percentage points are also observed under the Allen dataset standard.

These improvements confirm that the GL-Trans block is not merely an incremental modification but a decisive component that enhances long-range structure perception and mitigates boundary fragmentation and connectivity breaks that we observed in the baseline UKAN.

### 4.3.4. Summary

In summary, both quantitative metrics and qualitative results demonstrate that UKANFormer substantially outperforms mainstream segmentation models for coral reef mapping.

This improvement is not incidental, but arises from UKANFormer's targeted architectural refinement over UKAN. By integrating the GL-Trans block, it combines fine-grained boundary modeling with global semantic consistency—capabilities that conventional CNN or Transformer-only architectures often lack when trained with coarse labels.

These findings suggest that thoughtful architectural design can partially compensate for the limitations of noisy training labels, raising broader implications for model development in real-world, label-scarce environments.

### 4.4. Comparison between UKANFormer's Predictions and Allen Labels

To further evaluate the effectiveness of UKANFormer, we conduct a qualitative comparison between its predictions and the Allen Coral Atlas labels—one of the most widely used reference datasets in global coral reef mapping. **Figure 8** presents three representative satellite image patches, along with the corresponding Allen labels and UKANFormer outputs.

Visual inspection reveals that UKANFormer produces segmentation masks with greater spatial continuity and sharper boundary delineation, especially in transitional and edge regions. In contrast, the Allen labels often appear fragmented or overly conservative, potentially due to manual interpretation bias or coarse labeling strategies applied in ambiguous areas.

Notably, UKANFormer captures finer-scale, morphologically coherent coral reef structures, including low-contrast or partially occluded coral patches that are absent in the Allen labels. This advantage stems from its enhanced capacity to model complex spatial-spectral patterns via the KAN-augmented Transformer architecture.

Moreover, UKANFormer's predictions exhibit significantly improved connectivity across coral patches—a critical feature for downstream applications such as habitat assessment and conservation planning, where structural continuity is essential.

Importantly, these differences do not imply that the Allen labels are incorrect, but rather highlight the complementary value of automated segmentation models. While expert-drawn labels provide a strong baseline, models like UKANFormer can refine, extend, or scale up coral maps with greater spatial consistency.

In summary, this comparison underscores the practical utility of UKANFormer in enhancing coral segmentation under real-world, heterogeneous imaging conditions.

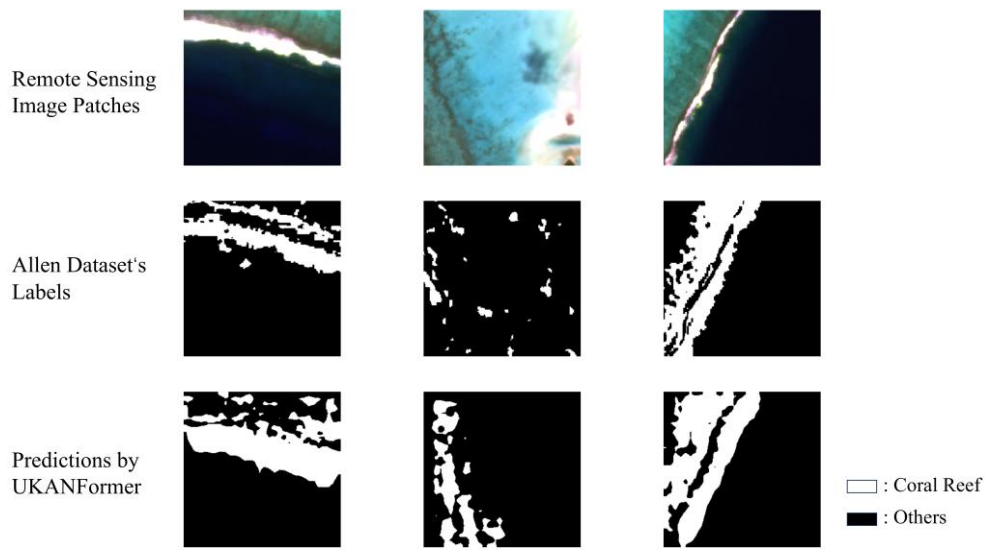

**Figure 8. Qualitative comparison between Allen Dataset labels and UKANFormer predictions.**
Visual comparison of coral reef detection results across three image patches. Top row: original satellite images; Middle row: Allen Dataset's labels; Bottom row: UKANFormer predictions. Our model demonstrates improved spatial coherence and boundary delineation compared to the Allen dataset's labels.

## 5. Discussion and Conclusion

In this work, we proposed UKANFormer, a novel semantic segmentation model that incorporates a Global-Local Transformer (GL-Trans) block to enhance feature representation. By training UKANFormer solely on the Allen dataset and evaluating it against expert-annotated ground truth data, we demonstrated that our model achieves significantly higher segmentation precision than the original dataset itself. This result not only highlights the potential of advanced transformer-based architectures in ecological remote sensing tasks but also reveals that well-designed models can surpass the limitations of their training data. UKANFormer offers not just a performance upgrade, but a model-level intervention for tackling the underexplored challenge of label-induced structural error—paving the way for more robust and scalable remote sensing frameworks under weak supervision.

The Allen Coral Atlas represents a groundbreaking achievement in global coral reef mapping, providing the first publicly available high-resolution maps of geomorphic zones and benthic substrates. By integrating data from multiple remote sensing platform and crowdsourced training labels, it laid a solid foundation for data-driven approaches in ecological remote sensing. However, the dataset's reliance on heuristic post-processing and limited validation introduced significant label noise, particularly in complex or transitional coral reef zones. Our findings demonstrate that UKANFormer, trained solely on this dataset, is capable of learning robust feature representations and achieves higher segmentation precision than the original labels when evaluated on expert-annotated ground truth. Similar observations have been reported in coral segmentation with sparse labels [41], as well as in weakly supervised and causal-invariance approaches in medical imaging[42,43], all suggesting that architectural advances can offset limitations in label quality.

This suggests that architectural improvements in model design can mitigate limitations in label quality. Our study challenges the prevailing assumption that better data is always a prerequisite for better performance, and instead points toward a more dynamic interplay between model structure and data quality. Moreover, this validates our central hypothesis: that a model architecture explicitly designed to compensate for structural inconsistencies in training labels—through enhanced global-local context modeling—can achieve high-quality predictions even in the absence of perfect supervision.

Despite the strong performance of UKANFormer, several limitations remain. Most notably, the model was trained solely on the Allen dataset, whose coarse labels limit the upper bound of achievable accuracy. Although UKANFormer surpasses the quality of its training labels in many regions, it does not yet exploit this capability to further improve the training data.

This suggests a promising avenue for future research: iterative refinement through model-guided self-training. By selectively incorporating high-confidence predictions from UKANFormer back into the training pipeline, it may be possible to progressively denoise and enhance the label quality, thereby achieving even higher segmentation pixel accuracy. In principle, this bootstrapping approach could lead to a virtuous cycle of label improvement and model enhancement, particularly in regions with inconsistent or ambiguous labels.

Additionally, incorporating auxiliary data sources (e.g., bathymetric models, water depth estimates) could further reduce uncertainty and improve generalizability. Ultimately, future work should aim not only to improve performance metrics, but to close the feedback loop between model predictions and dataset quality—enabling the recovery of coherent, ecologically meaningful coral structures that are often lost under current coarse labeling regimes. Such improvements are especially relevant from an ecological standpoint, as preserving connectivity among coral populations and understanding species-specific thermal resilience are key to forecasting coral reef survival under climate change[44,45].

Coarse and inconsistent labels remain a key bottleneck in global-scale coral segmentation, often resulting in fragmented, semantically weak maps that limit ecological interpretation. The Allen coral reef dataset marked a pioneering effort in global-scale coral reef mapping, offering a valuable foundation for the development of data-driven methods in marine remote sensing. However, its coarse labels impose limitations on segmentation pixel accuracy, particularly in complex coral reef environments.

Our study suggests a promising path toward more accurate and scalable coral reef habitat mapping, offering useful insights for both computer vision researchers and environmental scientists working to monitor and protect fragile coral reef ecosystems. In particular, it highlights the potential of deep learning models not only to interpret remote sensing data, but to critically evaluate and even improve upon existing large-scale environmental datasets.

**Acknowledgments:** This work was supported by the Key Research Project of the Technology Innovation Center for South China Sea Remote Sensing, Surveying, and Mapping Collaborative Application, Ministry of Natural Resources (grant number: RSSMCA-2024-A004), and the National Science Fund for Distinguished Young Scholars (grant number: 62425102). The numerical calculations in this paper have been done on the supercomputing system in the Supercomputing Center of Wuhan University. Special thanks are extended to Moorea IDEA at ETH Zürich for their invaluable contributions.